\documentclass[10pt,twocolumn,letterpaper]{article}

\usepackage{cvpr}
\usepackage{times}
\usepackage{epsfig}
\usepackage{graphicx}
\usepackage{amsmath}
\usepackage{amssymb}

\usepackage[numbers]{natbib}
\usepackage{subfiles}
\usepackage{adjustbox}
\usepackage{booktabs}
\usepackage{multirow}
\usepackage[capbesideposition=outside,capbesidesep=quad,capbesideposition={bottom,right}]{floatrow}
\usepackage{blindtext}
\usepackage{sidecap}
\usepackage{appendix}

\newcommand\paragraf[1]{\vskip 4px\noindent\textbf{#1}}
\newcommand\paragraff[1]{\noindent\textbf{#1}}
\newfloatcommand{capbtabbox}{table}[][\FBwidth]

\usepackage[pagebackref=true,breaklinks=true,letterpaper=true,colorlinks,bookmarks=false]{hyperref}

\cvprfinalcopy %

\def\cvprPaperID{****} %

\ifcvprfinal\fi
\begin{document}

\title{Learning to Generate 3D Training Data through Hybrid Gradient}

\author{Dawei Yang$^{1,2}$\\
$^1$University of Michigan\\
{\tt\small ydawei@umich.edu}
\and
Jia Deng$^2$\\
$^2$Princeton University\\
{\tt\small jiadeng@cs.princeton.edu}
}

\maketitle

\begin{abstract}
    Synthetic images rendered by graphics engines are a promising source for training deep networks. However, it is challenging to ensure that they can help train a network to perform well on real images, because a graphics-based generation pipeline requires numerous design decisions such as the selection of 3D shapes and the placement of the camera. In this work, we propose a new method that optimizes the generation of 3D training data based on what we call ``hybrid gradient''. We parametrize the design decisions as a real vector, and combine the approximate gradient and the analytical gradient to obtain the hybrid gradient of the network performance with respect to this vector. We evaluate our approach on the task of estimating surface normal, depth or intrinsic decomposition from a single image. Experiments on standard benchmarks show that our approach can outperform the prior state of the art on optimizing the generation of 3D training data, particularly in terms of computational efficiency.
    \end{abstract}

\begin{figure*}[tb]
    \centering
    \includegraphics[width=0.9\linewidth]{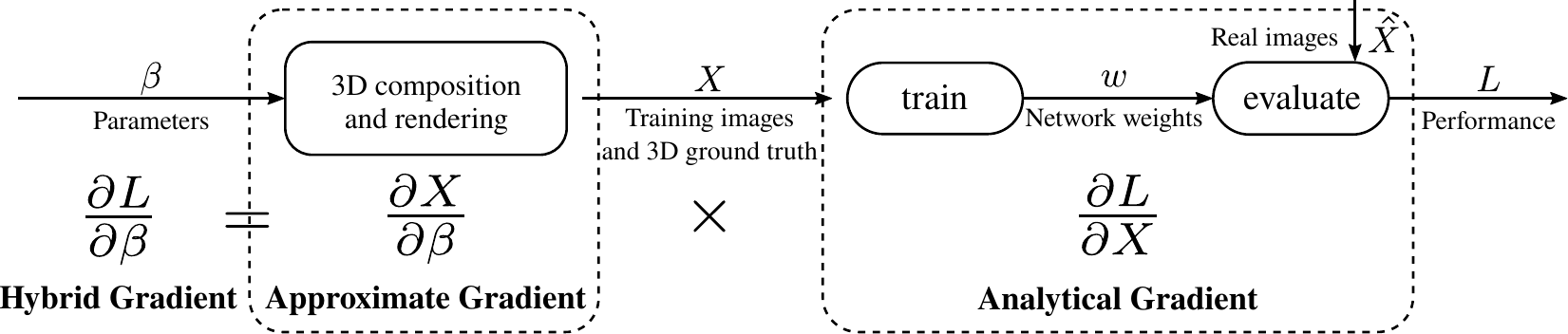}
    \caption{Our hybrid gradient method. We parametrize the design decisions as a real vector $\beta$ and optimize the function of performance $L$ with respect to $\beta$. From $\beta$ to the generated training images and ground truth, we compute the approximate gradient by averaging finite difference approximations. From training samples $X$ to $L$, we compute the analytical gradient through backpropagation with unrolled training steps.}
    \label{fig:pull_figure}
\end{figure*}
\section{Introduction}

Synthetic images rendered by graphics engines have emerged as a promising source of training data for deep networks, especially for vision and robotics tasks that involve perceiving 3D structures from RGB pixels~\citep{ButlerECCV2012MPI-Sintel,YehTOG2012Synthesizing,VarolCVPR2017SURREAL,RosCVPR2016SYNTHIA,McCormacICCV2017SceneNet,XiaCVPR2018Gibson,ChangICCV2017Matterport3D,KolveArXiv2017AI2THOR,SongCVPR2017Semantic,RichterECCV2016Playing,RichterICCV2017Playing,ZhangCVPR2016Physically,LiECCV2018CGIntrinsics}. A major appeal of generating training images from computer graphics is that they have a virtually unlimited supply and come with high-quality 3D ground truth for free. 

Despite its great promise, however, using synthetic training images from graphics poses its own challenges. One of them is ensuring that the synthetic training images are useful for real-world tasks, in the sense that they help train a network to perform well on real images. Ensuring this is challenging because a graphics-based generation pipeline requires numerous design decisions, including the selection of 3D shapes, the composition of scene layout, the application of texture, the configuration of lighting, and the placement of the camera. These design decisions can profoundly impact the usefulness of the generated training data, but have largely been made manually by researchers in prior work, potentially leading to suboptimal results. 

In this paper, we address the problem of automatically optimizing a generation pipeline of synthetic 3D training data, with the explicit objective of improving the generalization performance of a trained deep network on real images. %

One idea is black-box optimization: we try a particular configuration of the pipeline, use the pipeline to generate training images, train a deep network on these images, and evaluate the network on a validation set of real images. We can treat the performance of the trained network as a black-box function of the configuration of the generation pipeline, and apply black-box optimization techniques. Recent works \citep{YangCVPR2018Shape,Ruiz2018Simulate} have explored this exact direction. \citet{YangCVPR2018Shape} use genetic algorithms to optimize the 3D shapes used in the generation pipeline. In particular, they start with a collection of simple primitive shapes such as cubes and spheres, and evolve them through mutation and combination into complex shapes, whose fitness is determined by the generalization performance of a trained network. They show that the 3D shapes evolved from scratch can provide more useful training data than manually created 3D CAD models.
Meanwhile, \citet{Ruiz2018Simulate} use black box reinforcement learning algorithms to optimize the parameters of a simulator, and shows that their approaches converge to the optimal solution in controlled experiments and can indeed discover good sets of parameters.

The advantage of black-box optimization is that it assumes nothing about the function being optimized as long as it can be evaluated. As a result, it can be applied to any existing function, including advanced photorealistic renderers. On the other hand, black-box optimization is computationally expensive---knowing nothing else about the function, it needs many trials to find a reasonable update to the current solution. In contrast, gradient-based optimization can be much more efficient by assuming the availability of the analytical gradient, which can be efficiently computed and directly correspond to good updates to the current solution, but the downside is that the analytical gradient is often unavailable, especially for many advanced photorealistic renderers. 

In this work, we propose a new method that optimizes the generation of 3D training data based on what we call ``hybrid gradient''. The basic idea is to make use of the analytical gradient where they are available, and combine them with black-box optimization for the rest of the function. We hypothesize that hybrid gradient will lead to more efficient optimization than black-box methods because it makes use of the partially available analytical gradient. 

Concretely, if we parametrize the design decisions as a real vector $\beta$, the function mapping $\beta$ to the network performance $L$ can decompose into two parts: (1) from the design parameters $\beta$ to the generated training images $X$, and (2) from the training images $X$ to the network performance $L$. The first part often does not have analytical gradient, due to the use of advanced photorealistic renderers. We instead compute the approximate gradient by averaging finite difference approximations along random directions~\citep{HoriaNIPS2018RandomSearch}. For the second part, we compute the analytical gradient through backpropagation---with SGD training unrolled, the performance of the network is a differentiable function of the training images. Then we combine the approximate gradient and the analytical gradient to obtain the hybrid gradient of the network performance $L$ with respect to the parameters $\beta$, as illustrated in Fig.~\ref{fig:pull_figure}. 

A key ingredient of our approach is representing design decisions as real vectors of fixed dimensions, including the selection and composition of shapes. \citet{YangCVPR2018Shape} represent 3D shapes as a finite set of graphs, one for each shape. This representation is suitable for a genetic algorithm but is incompatible with our method. Instead, we propose to represent 3D shapes as random samples generated by a Probabilistic Context-Free Grammar (PCFG)~\citep{HarrisonBook1978Introduction}. To sample a 3D shape, we start with an initial shape, and repeatedly sample a production rule in the grammar to modify it. The (conditional) probabilities of applying the production rules are parametrized as a real vector of a fixed dimension. 

Our approach is novel in multiple aspects. First, to the best of our knowledge, we are the first to propose the idea of hybrid gradient,  i.e.\@ combining approximate gradient and analytical gradient, especially in the context of optimizing the generation of 3D training data. Second, we propose a novel integration of PCFG-based shape generation and our hybrid gradient approach.  

We evaluate our approach on the task of estimating surface normal, depth and intrinsic components from a single image. Experiments on standard benchmarks and controlled settings show that our approach can outperform the prior state of the art on optimizing the generation of 3D training data, particularly in terms of computational efficiency.

\section{Related Work}
\paragraff{Generating 3D training data}
Synthetic images generated by computer graphics have been extensively used for training deep networks for numerous tasks,
including single image 3D reconstruction~\citep{SongCVPR2015SUN,Hua3DV2016SceneNN,McCormacICCV2017SceneNet,JanochICCV2011Category,YangCVPR2018Shape,Change2015ShapeNet},
optical flow estimation~\citep{MayerIJCV2018GoodSynthetic,ButlerECCV2012MPI-Sintel,GaidonCVPR2016VirtualKITTI}, human pose estimation~\citep{VarolCVPR2017SURREAL,Chen3DVDeep3DPose}, action recognition~\citep{SouzaCVPR2017Procedural}, visual question answering~\citep{JohnsonCVPR2017CLEVR}, and many others~\citep{QiuACMMM2017UnrealCV,Martinez-GonzalezArXiv2018UnrealROX,XiaCVPR2018Gibson,TobinArXiv2017Domain,RichterICCV2017Playing,RichterECCV2016Playing,WuICLRw2018House3D}. The success of these works has demonstrated the effectiveness of synthetic images. 

To ensure the relevance of the generated training data to real-world tasks, a large amount of manual effort has been necessary, particularly in acquiring 3D assets such as shapes and scenes~\citep{Change2015ShapeNet,JanochICCV2011Category,ChoiArXiv2016Scan,XiangECCV2016ObjectNet3D,Hua3DV2016SceneNN,McCormacICCV2017SceneNet,SongCVPR2017Semantic}. 
To reduce manual labor, some heuristics have been proposed to generate 3D configurations automatically.
For example, \citet{ZhangCVPR2016Physically} design an approach to use the entropy of object masks and color distribution of the rendered images to select sampled camera poses. ~\citet{McCormacICCV2017SceneNet} simulate gravity for physically plausible object configurations inside a room.

Apart from simple heuristics, prior work has also performed automatic optimization of 3D configurations towards an explicit objective.  %
For example, \citet{YehTOG2012Synthesizing} synthesize layouts with the target of satisfying constraints such as non-overlapping and occupation.
\citet{JiangIJCV2018SceneSynthesis} learn a probabilistic grammar model for indoor scene generation, with parameters learned using maximum likelihood estimation on the existing 3D configurations in SUNCG~\citep{SongCVPR2017Semantic}.
Similarly, \citet{VeeravasarapuArXiv2017SceneGeneration} tune the parameters for stochastic scene generation using generative adversarial networks, targeting at making synthetic images indistinguishable from real images.
\citet{QiCVPR2018Human-Centric} synthesize 3D room layouts based on human-centric relations among furniture, to achieve visual realism, functionality and naturalness of the scenes. However, these optimization objectives are different from ours, which is the generalization performance of a trained network on real images. %

In terms of generating 3D training data, the closest prior works to ours are those of \citep{YangCVPR2018Shape,kar2019metasim,Ruiz2018Simulate}. Specifically, \citet{YangCVPR2018Shape} use a genetic algorithm to optimize the 3D shapes used for rendering synthetic training images. Their optimization objective is the same as ours except that their optimization method is different: they leverage evolution-based approach as apposed to using gradient information. Similarly, Meta-Sim~\citep{kar2019metasim} also tries to optimize 3D parameters with REINFORCE towards better task generalization performance, and \citet{Ruiz2018Simulate} learn a policy for simulator parameters also using REINFORCE. However, they do not backpropagate analytical gradient from the meta-objective, so their algorithms can be considered as black-box estimation by multiple trials, with an improved efficient sampling strategy (REINFORCE). In our experiments, we compared to an algorithm that has been shown competitive to REINFORCE in training deep policy networks~\cite{HoriaNIPS2018RandomSearch,SalimansArXiv2017evolution,Song2020ES-MAML}.

\paragraf{Unrolling and backpropagating through network training}
One component of our approach is unrolling and backpropagating through the training iterations of a deep network. This is a technique that has often been used by existing work in other contexts, including hyperparameter optimization~\citep{MaclaurinICML2015Gradient}, meta-learning~\citep{AndrychowiczNIPS2016Learning,HaICLR2017Hypernetworks,MunkhdalaiICML2017MetaNetworks,LiICLR2017Learning,FinnNIPS2018MAML} and others~\citep{crfasrnn_iccv2015,ChenICCV2015Learning}.
Our work is different in that we apply this technique in a novel context: it is used to optimize the generation of 3D training data, and the gradient with respect to the input images is integrated with approximate gradient to form hybrid gradient.

\paragraf{Hyperparameter optimization}
Our method is connected to hyperparameter optimization in the sense that we can treat the design decisions of the 3D generation pipeline as hyperparameters of the training procedure.

Hyperparameter optimization of deep networks is typically approached as black-box optimization~\citep{BergstraJMLR2012RandomSearch,Bergstra2011Algorithms,LacosteUAI2014Ensemble,BrochuArXiv2010Bayesian}. While \citet{Klatzer2015ContinuousHL} propose a bi-level gradient-based approach for continuous hyperparameter optimization of Support Vector Machines, but it has not been applied to deep networks and 3D generation.
Since black-box optimization does not make assumption of the function being optimized, it requires repeated evaluation of the function, which is expensive in this case because it contains the process of training and evaluating a deep network.
In contrast, we combine the analytical gradient from backpropagation and the approximate gradient from generalized finite difference for more efficient optimization.  

\paragraf{Domain Adaptation}
Researchers have also applied domain adaptation techniques to transfer the knowledge learned from synthetic data to real data.
Like domain adaptation, our method involves data from two domains: synthetic and real. However, our setting is different: in domain adaptation, the distribution of training data is fixed; in our setting, we are concerned about \emph{generating and changing} the distribution of training data in the source domain.
\paragraf{Differentiable Rendering} Researchers have also explored differentiable rendering engines to obtain the gradient with respect to the input 3D content such as mesh vertices, lighting intensity \etc~\citep{Loper2014OpenDR,Kato2018Renderer,Wu2017SceneDeRendering,Li2018MonteCarlo,Che2018Transport}. Generally, they obtain the gradient through backpropagation~\citep{Kato2018Renderer,Loper2014OpenDR} or sampling~\citep{Wu2017SceneDeRendering,Li2018MonteCarlo,Che2018Transport}. The differentiable renderers often assume simple surface reflectance and illumination model, and they are typically developed for a specific 3D input format (such as triangle meshes and directional lighting) or a specific rendering algorithm (such as path tracing). In fact, we are not aware of any photorealistic differentiable renderer that is differentiable over a shape parametrization that allows not only continuous deformation but also topology change. In our method, we assume nothing about the rendering engine and obtain the gradient with respect to the decision vector by approximation, bypassing the surface and illumination model or any rendering algorithms. So our method is flexible and not limited by choices of graphics engines of any kind.

\section{Problem Setup}

Suppose we have a probabilistic generative pipeline. We use a deterministic function, $f(\beta, r)$ to represent the sampling operation. This function $f$ takes the real vector $\beta$ and the random seed $r$ as input.
An image and its 3D ground truth are computed by evaluating the function $f(\beta, r)$.
By choosing $n$ different random seeds $r$, we obtain a dataset of size $n$ for training:
\begin{equation}
X = (f(\beta, r^{(1)}), f(\beta, r^{(2)}), \cdots, f(\beta, r^{(n)}))
\end{equation}
Then, a deep neural network with initialized weights $w_0$ is trained on the training data $X$, with the function $\mathrm{train}(w_0, X)$ representing the optimization process and generating the weights of the trained network.

The network is then evaluated on real data $\hat X$ with a validation loss $l_\mathrm{eval}$ to obtain a generalization performance $L$:
\begin{equation}
    L = l_\mathrm{eval}(\mathrm{train}(w_0, X), \hat X)
\end{equation}

Combining the above two functions, $L$ is a function of $\beta$,
and the task is to optimize this value $L$ with respect to the parameters $\beta$.

As we mentioned in the previous section, black-box algorithms typically need repetitive evaluations of this function, which is expensive.

\begin{figure*}[tb]
  \centering
  \includegraphics{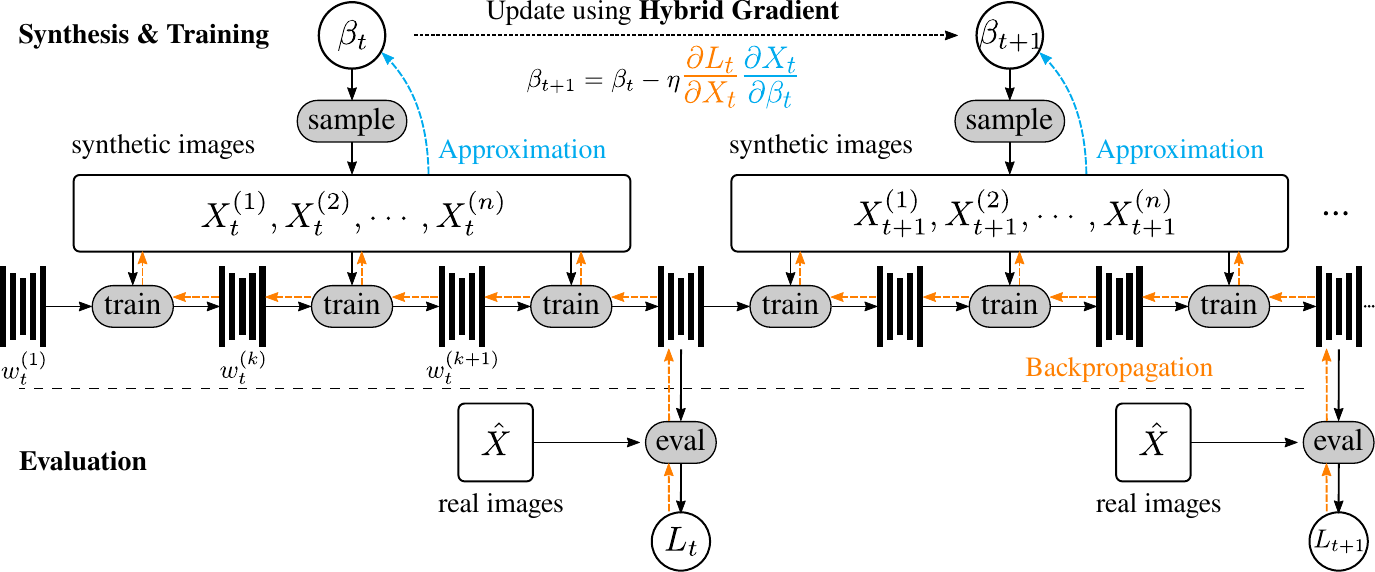}
  \caption{The details of using ``hybrid gradient'' to incrementally update $\beta$ and train the network. The analytical gradient is computed by backpropagating through unrolled training steps (colored in orange).
  The numerical gradient is computed using finite difference approximation by sampling in a neighborhood of $\beta_t$ (colored in cyan). Then $\beta_t$ is updated using hybrid gradient, and the trained network weights are retained for the next timestamp $t+1$.
  }
  \label{fig:approach}
\end{figure*}

\section{Approach}
\subsection{Generative Modeling of Synthetic Training Data}
We decompose the function $f(\beta, r)$ into two parts:
3D composition and rendering.

\paragraf{3D composition}
Context-free grammars have been used in scene generation~\citep{JiangIJCV2018SceneSynthesis,QiCVPR2018Human-Centric} and in the parsing of the Constructive Solid Geometry (CSG) shapes~\citep{SharmaCVPR2018CSGNet} because they can represent shapes and scenes in a flexible and composable manner.
Here, we design a probabilistic context-free grammar (PCFG)~\citep{HarrisonBook1978Introduction} to control the random generation of unlimited shapes~\citep{FoleyBook1990CSG}.

In a PCFG, a tree is randomly sampled given a set of probabilities.
Starting from a root node, the nodes are expanded by randomly sampling probabilistic rules repeatedly until all the leaf nodes cannot expand.
Since multiple rules may apply, the parameters in a PCFG define the probability distribution of applying different rules.

In our PCFG, a shape is constructed by composing two other shapes through union and difference;
this construction is recursively applied until all leaf nodes are a predefined set of concrete primitive shapes (terminals). The parameters include the parameters of primitive shapes as well as the probability of either expanding the node or replacing it with a terminal.

Given our PCFG model with the probability parameters $\beta_S$, a 3D shape $S$ can be composed by computing a deterministic function $f_\mathcal{S}$ given $\beta_S$ and a random string $r_S$ as the input:
\begin{equation}\label{eq:sample_instance}
   S = f_\mathcal{S}(\beta_S, r_S)
\end{equation}

\paragraf{Rendering training images} we use a graphics renderer $R$ to render the composed shape $S$.
The rendering configurations $P$ (e.g.\@ camera poses), are also sampled from a distribution controlled by a set of parameters $\beta_R$ (with a random string $r_R$):
\begin{equation}\label{eq:sample_render}
   P = f_\mathcal{R}(\beta_R, r_R)
\end{equation}

Now that we have Eq.~\ref{eq:sample_instance} and \ref{eq:sample_render},
 The full function for training data generation can be represented as follows: 
\begin{equation}\label{eq:render_formulation}
  f(\beta, r) = R(S, P) =  R(f_\mathcal S(\beta_S, r_S), f_\mathcal R(\beta_R, r_R))
\end{equation}
where $\beta=(\beta_R, \beta_S)$ and $r = (r_R, r_S)$.

By sampling different random strings $r$, we obtain a set of training images and their 3D ground truth $X$.

\subsection{Hybrid Gradient}
After training deep network on synthetic training data $X$,
the network is evaluated on a set of validation images $\hat X$ to obtain the generalization loss $L$.

Recall that to compute the hybrid gradient $\frac{\partial L}{\partial \beta}$ to optimize $\beta$, 
we multiply two types of gradient: the gradient of network training $\frac{\partial L_t}{\partial X}$ and the gradient of image generation $\frac{\partial X}{\partial\beta}$, as is shown in Fig.~\ref{fig:approach}.

\paragraf{Analytical gradient from backpropagation}
We assume the network is trained on a set of previously generated training images $X^{(1)},X^{(2)},\cdots,X^{(n)}$. %
Without loss of generality, we assume mini-batch stochastic gradient descent (SGD) with a batch size of 1 is used for weight update.
Let function $g$ denote the SGD step and let $l_\mathrm{train}$ denote the training loss:
{\small
\begin{equation}\label{eq:sgd_forward}
\begin{split}
w^{(k+1)} &= w^{(k)} - \eta\frac{\partial l_\mathrm{train}(w^{(k)}, X^{(k)})}{\partial w^{(k)}} 
\\ &= g(w^{(k)}, X^{(k)}; l_\mathrm{train},\eta)
\end{split}
\end{equation}
}
Note that the SGD step $g$ is differentiable with respect to the network weights $w^{(k)}$ as well as the training batch $X^{(k)}$,
if our training loss $l_\mathrm{train}$ is twice (sub-)differentiable.
This requirement is satisfied in most practical cases.
To simplify the equation, we assume the training loss $l_\mathrm{train}$ and the learning rate $\eta$ do not change during one update step of $\beta$,
so the variables can be safely discarded in the equation.

Therefore, the gradient from the generalization loss $L$ to each sample $X^{(k)}$ can be computed through backpropagation. Given Eq.~\ref{eq:sgd_forward}:
\begin{equation}
\small\begin{split}
&\frac{\partial{L}}{\partial X^{(k)}} = \frac{\partial L}{\partial w^{(k+1)}}\cdot\frac{\partial w^{(k+1)}}{\partial X^{(k)}}
 = \frac{\partial L}{\partial w^{(k+1)}}\cdot g'_2(w^{(k)}, X^{(k)}) \\
&\frac{\partial{L}}{\partial w^{(k)}} = \frac{\partial{L}}{\partial w^{(k+1)}}\cdot\frac{\partial w^{(k+1)}}{\partial w^{(k)}}
= \frac{\partial{L}}{\partial w^{(k+1)}}\cdot g'_1(w^{(k)}, X^{(k)})
\end{split}
\end{equation}
with the initial value $\frac{\partial L}{\partial w^{(n+1)}}$ computed from the validation loss $l_\mathrm{eval}$:
\begin{equation}
\frac{\partial{L}}{\partial{w}^{(n+1)}} = l'_\mathrm{eval}(w^{(k+1)}, \hat X)
\end{equation}

\paragraf{Approximate gradient from finite difference}
For the formulation in Eq.~\ref{eq:render_formulation}, the graphics renderer can be a general black box and non-differentiable.
We can approximate the gradient of each rendered image with ground truth $X^{(1)}, X^{(2)}, \cdots$ with respect to the generation parameters $\beta$ using generalized finite difference. We adopt the form of \citep{HoriaNIPS2018RandomSearch} because this gradient approximation algorithm in Random Search has been shown effective for training deep policy networks~\citep{HoriaNIPS2018RandomSearch,SalimansArXiv2017evolution,Song2020ES-MAML}.
Concretely, we sample a set of noise from an uncorrelated multivariate Gaussian distribution:
\begin{equation}
\delta_1,\delta_2,\cdots,\delta_m \sim \mathcal{N}(\mathbf{0}, \sigma I)
\end{equation}
Next, we approximate the Jacobian for each sample ($\otimes$ denotes outer product):
\begin{equation}\label{eq:jacobian}
\frac{\partial X^{(i)}}{\partial\beta} \approx
\frac1m\sum_{j=1}^m
\frac{f_\mathcal D(\beta+\delta_j, r_i)-f_\mathcal D(\beta-\delta_j,r_i)}{2\|\delta_j\|} \otimes \frac{\delta_j}{\|\delta_j\|}
\end{equation}

\paragraf{Incremental training}
Following \citet{YangCVPR2018Shape}, we incrementally train the network $w$ along with the update of $\beta$, instead of initializing $w^{(1)}$ from scratch each time.
At timestamp $t$, we update $\beta_t$ with the hybrid gradient;
for network weights, we keep the trained network in timestamp $t$ for initialization in the next timestamp $t+1$:
\begin{equation}
\small\begin{split}
&\beta_{t+1} = \beta_t - \gamma\frac{\partial L_t}{\partial\beta_t} = \beta_t - \gamma\sum_{i=1}^{n}
\frac{\partial L_t}{\partial X_t^{(i)}}\cdot\frac{\partial X_t^{(i)}}{\partial\beta_t} \\
& w_{t+1}^{(1)} = w_t^{(n+1)}
\end{split}
\end{equation}

\section{Experiments}
\paragraf{Datasets}
We evaluate our algorithm on four different datasets, and three standard prediction tasks for single-image 3D. The input is an RGB image and the output is pixel-wise surface normal, depth, or albedo shading map.

Specifically, we experiment on the task of surface normal estimation on two real datasets: MIT-Berkeley Intrinsic Images Dataset (MBII)~\citep{BarronTPAMI2015Shape}, which focuses on images of single objects and NYU Depth~\citep{SilbermanECCV2012Indoor}, which focuses on indoor scenes.
For the other two datasets, we illustrate that our method can easily extend to other 3D setups. We experiment on the task of depth estimation on the renderings of the scanned human faces in the Basel Face Model dataset~\citep{Paysan2009}, and on the task of intrinsic image decomposition and evaluate on the renderings of ShapeNet~\citep{Change2015ShapeNet} shapes.

\paragraf{Baselines}
For comparison, we implemented a black-box optimization method.
Random search~\citep{Baba1981Random} has been extensively explored~\citep{Flaxman2005Bandit,Nesterov2017Random,HoriaNIPS2018RandomSearch} as a derivative-free optimization method, and \citet{HoriaNIPS2018RandomSearch} have shown that their simple version, Basic Random Search, has comparable performance compared to typical reinforcement learning algorithms.
Therefore, we re-implemented their Basic Random Search such that this baseline has the same setting
as in our method, while the only difference is that the gradient from the validation loss is obtained through sampling instead of hybrid gradient.
We also compare against baselines with a random $\beta$ baseline in the following experiments. In these baselines, the networks are trained on a dataset generated using multiple random but fixed $\beta$, and the weight snapshots with the best validation performance are used to evaluate on the test set.

These two baselines, along with our hybrid gradient method, all use information from the validation set but in a different way: hybrid gradient backpropagates the gradient of the validation performance to update $\beta$; random search samples $\beta$ to get the gradient from the validation performance; the random $\beta$ baseline fixes the dataset and uses the validation performance to select the best network snapshot.

In all of our experiments, the network weights are updated using only synthetic images in the training iterations, and the generalization loss is computed only on the validation split of the datasets mentioned above. The decision vector $\beta$ is updated using RMSprop~\citep{Tieleman2012} for hybrid gradient.

For MBII, we use pure synthetic shapes~\citep{YangCVPR2018Shape} to render training images.
We first compare our method with ablation baselines, then show that our algorithm is better than the previous state of the art on MBII.
For NYU Depth, we base our generative model on SUNCG~\citep{SongCVPR2017Semantic} and augment the original 3D configurations in \citet{ZhangCVPR2016Physically}.
For Basel Face Model, we sample synthetic faces from a morphable model and evaluate on the renderings of scanned faces. For the intrinsic image decomposition task, we sample textures from a simple procedural pipeline and attach the synthetic textures to SUNCG shapes~\citep{SongCVPR2017Semantic}, and evaluate on renderings of ShapeNet shapes~\citep{Change2015ShapeNet}.

\subsection{Normal Estimation on MIT-Berkeley Intrinsic Images}

\begin{table*}[bt]
  \floatbox[\capbeside]{table}[.55\textwidth]{
  \caption{Ablation Study: the diagnostic experiment to compare with random but fixed $\beta$. We sample $10$ values of $\beta$ in advance, and then train the networks with the same setting as in hybrid gradient.
  The best, median and worst performance is reported on the test images, and the corresponding values of $\beta$ are used to initialize $\beta_0$ for hybrid gradient for comparison. The results show that our approach is consistently better than the baselines with fixed $\beta$.}
  \label{table:mbii_ablation}}
  {
  \begin{adjustbox}{max width=\linewidth}
  \begin{tabular}{clcccccc}
  
    \toprule
    & & \multicolumn{3}{c}{Summary Stats $\uparrow$} & \multicolumn{3}{c}{Errors $\downarrow$} \\
    \cmidrule(r){3-5} \cmidrule(l){6-8}
    & & $\leq 11.25^\circ$ & $\leq 22.5^\circ$ & $\leq 30^\circ$ & MAE & Median & MSE \\
    \midrule
    \multirow{3}{*}{Fixed $\beta$}
    & $\beta=\beta_\mathrm{best}$   & $19.9\%$ & $52.7\%$ & $70.5\%$ & $24.0^\circ$ & $21.5^\circ$ & $0.2282$ \\
    & $\beta=\beta_\mathrm{median}$ & $20.7\%$ & $50.9\%$ & $67.5\%$ & $24.8^\circ$ & $22.1^\circ$ & $0.2461$ \\
    & $\beta=\beta_\mathrm{worst}$  & $17.9\%$ & $46.7\%$ & $64.6\%$ & $25.6^\circ$ & $23.8^\circ$ & $0.2553$ \\
    \midrule
    \multirow{3}{*}{Hybrid gradient}
    & $\beta_0=\beta_\mathrm{best} $ & $22.7\%$ & $58.5\%$ & $73.9\%$ & $22.5^\circ$ & $19.3^\circ$ & $0.2065$ \\
    & $\beta_0=\beta_\mathrm{median} $ & $24.0\%$ & $60.1\%$ & $75.7\%$ & $21.8^\circ$ & $18.8^\circ$ & $0.1938$ \\
    & $\beta_0=\beta_\mathrm{worst} $ & $26.0\%$ & $58.6\%$ & $73.9\%$ & $22.0^\circ$ & $19.1^\circ$ & $0.1998$ \\
    \bottomrule
  \end{tabular}
  \end{adjustbox}}
\end{table*}

\begin{table*}[hbt]
  \floatbox[\capbeside]{table}[.55\textwidth]{
  \caption{Our approach compared to previous work, on the test set of MIT-Berkeley images~\citep{BarronTPAMI2015Shape}. The results show that our approach is better than the state of the art as reported in \cite{YangCVPR2018Shape}.\vspace{15px}}
  \label{table:mbii_best}
  }
   {
  \begin{adjustbox}{max width=.9\linewidth}
  \begin{tabular}{lcccccc}
    \toprule
    & \multicolumn{3}{c}{Summary Stats $\uparrow$} & \multicolumn{3}{c}{Errors $\downarrow$} \\
    \cmidrule(r){2-4} \cmidrule(l){5-7}
    & $\leq 11.25^\circ$ & $\leq 22.5^\circ$ & $\leq 30^\circ$ & MAE & Median & MSE \\
    \midrule
    SIRFS~\citep{BarronTPAMI2015Shape} &  $20.4\%$ & $53.3\%$ & $70.9\%$ & $26.2^\circ$ & --- & $0.2964$ \\
    Evolution~\citep{YangCVPR2018Shape}(Reported) & $21.6\%$ & $ 55.5\% $ & $73.5\%$ & $23.3^\circ$ & --- & $0.2204$ \\
    Evolution~\citep{YangCVPR2018Shape}(Our Impl.) & $23.0\%$ & $ 58.3\%$ & $73.8\%$ & $22.5^\circ$ & $\mathbf{18.8}^\circ$ & $0.2042$ \\
    \midrule
    Basic Random Search~\citep{HoriaNIPS2018RandomSearch} & $21.9\%$ & $\mathbf{59.6\%}$ & $74.0\%$ & $22.8^\circ$ & $19.2^\circ$ & $0.2106$ \\
    Hybrid gradient & $\mathbf{24.5\%}$ & $59.3\%$ & $\mathbf{74.3\%}$ & $\mathbf{22.0^\circ}$ & $18.9^\circ$ & $\mathbf{0.1984}$ \\
    \bottomrule
  \end{tabular}
\end{adjustbox}}
\end{table*}

Following the work of \citet{YangCVPR2018Shape}, we recover the surface normals of an object from a single image.

\paragraf{Synthetic shape generation}
In \citet{YangCVPR2018Shape}, a population of primitive shapes such as cylinders, spheres and cubes are evolved and 
rendered to train deep networks.
The evolution operators include transformations of individual shapes and the boolean operations of shapes in Constructive Solid Geometry (CSG)~\citep{FoleyBook1990CSG}.
In our algorithm, we also use the CSG grammar for our PCFG:
{\scriptsize
\begin{verbatim}
    S => E
    E => C(E, T(E)) | P
    C => union | subtract
    P => sphere | cube | truncated_cone | tetrahedron
    T => attach * rand_transl * rand_rotate * rand_scale
\end{verbatim}
}

\begin{figure}[b]
  \centering
  \includegraphics[width=\linewidth]{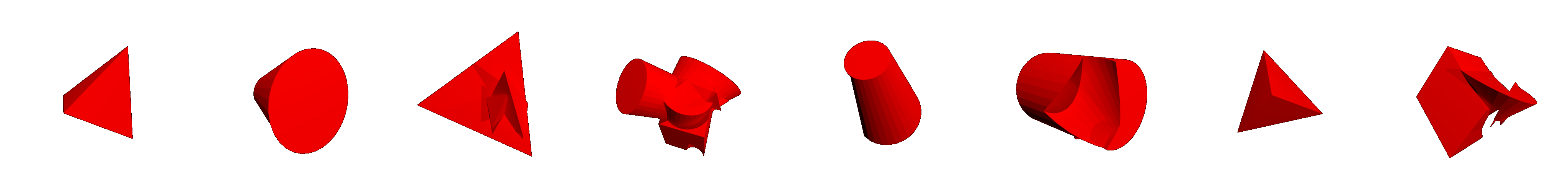}
  \caption{Sampled shapes from our probabilistic context-free grammar, with parameters optimized using hybrid gradient.}
  \label{fig:csg_shapes}
\end{figure}

In this PCFG, the final shape \verb+S+ is generated by recursively composing (\verb+C+) other shapes \verb+E+ with transformations \verb+T+, until primitives \verb+P+ are sampled at all \verb+E+ nodes.
The parameter vector $\beta$ consists of three parts:
(1) The probability of the different rules;
(2) The means and variations of log-normal distributions controlling shape primitives (\verb+P+), such as
the radius of the sphere;
(3) The means and variations of log-normal distributions controlling transformation parameters (\verb+T+),
such as scale values.
Examples of sampled shapes are shown in Fig.~\ref{fig:csg_shapes}. For the generalization loss $L$, we compute the mean angle error of predictions on the training set of the MIT-Berkeley dataset.

\paragraf{Training setup}
For network training and evaluation, we follow \citet{YangCVPR2018Shape} and train the Stacked Hourglass Network~\citep{NewellECCV2016Stacked} on the images, and use the standard split of the MBII dataset for the optimization of $\beta$ and testing.

We report the performance of surface normal directions with the metrics commonly used in previous works, including mean angle error (MAE), median angle error, mean squared error (MSE), and the proportion of pixels that normals fall in an error range ($\le N^\circ$). See Appendix A for detailed definitions.

\paragraf{Ablation study} We first sample 10 random values of $\beta$ and fix those values in advance. Then, for each $\beta$, we sample 3D shapes and render images to train a network, with the same training and evaluation configurations as in our hybrid gradient, except that we do not update $\beta$.
We then report the best, median and worst performance of those 10 networks, and label the corresponding $\beta$ as $\beta_\mathrm{best}$, $\beta_\mathrm{median}$ and $\beta_\mathrm{worst}$.
In hybrid gradient, we then initialize $\beta_0$ from these three values, run our algorithm, and report the performance on test images also in Table~\ref{table:mbii_ablation}.

From the table we can observe that training with a fixed $\beta$ can hardly match the performance of our method, even with multiple trials.
Instead, our hybrid gradient approach can optimize $\beta$ to a reasonable performance
regardless of different initialization ($\beta_\mathrm{best/median/worst}$).
This simple diagnostic experiment demonstrates that our algorithm is working correctly: the optimization of $\beta$ is necessary in order to generate useful synthetic images for training networks.

\begin{figure}
  \centering
  \includegraphics[width=\linewidth]{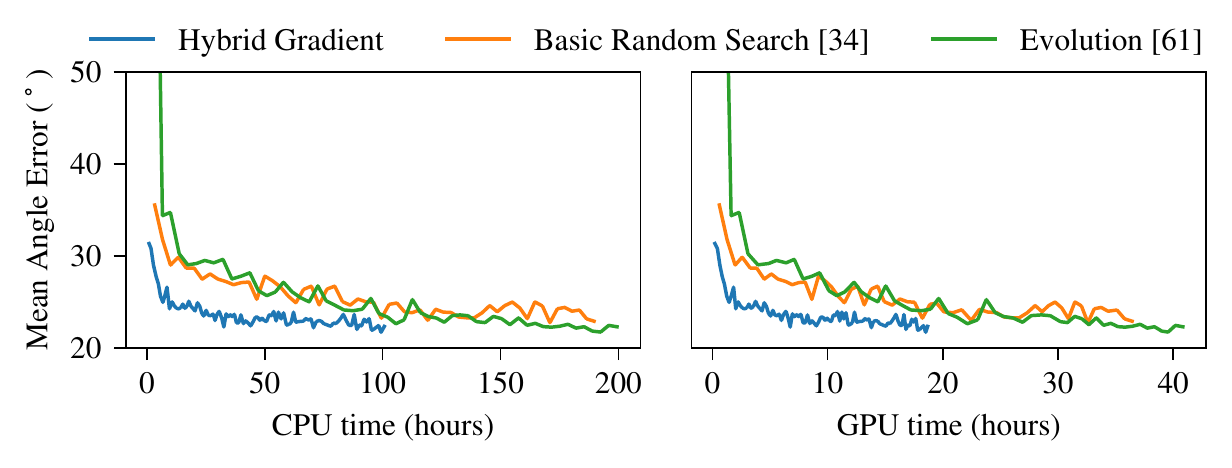}
  \caption{Mean angle error on the test images \vs computation time, compared to two black-box optimization baselines.}
  \label{fig:speed_curve}
\end{figure}

\paragraf{Comparison with the state of the art}
In addition to Basic Random Search as mentioned earlier,
in this experiment we also compare with \citet{YangCVPR2018Shape}, a state-of-the-art method on MIT-Berkeley Intrinsic Images.

In Shape Evolution~\citep{YangCVPR2018Shape}, a population of shapes are evolved, and fitness scores for individual shapes are computed using a network trained on an incremental dataset and evaluated on the validation set.
We compose our shapes in mesh representations, slightly different from the implicit functions in \citet{YangCVPR2018Shape}. Therefore, we re-implemented their algorithm with mesh representations for a fair comparison.
We follow \citet{YangCVPR2018Shape} for the initialization of $\beta$, and train the networks and update $\beta$ for the same number of steps. We then report the test performance of the network that has the best validation performance.
The results are shown in Table~\ref{table:mbii_best}.

We also run the experiments on the same set of CPUs and GPUs, sum the computation time, and plot the mean angle error (on the test set) with respect to the CPU time and GPU time  Fig.~\ref{fig:speed_curve}). We see that our algorithm is more efficient than the above baselines. This is natural, because when computing $\partial L/\partial \beta^{(t)}$ in black-box algorithms, for each sample of $\beta^{(t)}+\delta_j$, one needs to train one network to evaluate the performance $L$, while in hybrid gradient, only a forward training pass and a backpropagation pass for a single network are required to compute $\partial L/\partial X$. Shapes sampled from our optimized PCFG are shown in Fig.~\ref{fig:csg_shapes}.

\subsection{Normal Estimation on NYU Depth}
\paragraff{Scene perturbation}
We design our scene generation grammar as an augmentation of collected SUNCG scenes~\citep{SongCVPR2015SUN}
with the cameras from \citet{ZhangCVPR2016Physically}:
{\scriptsize
\begin{verbatim}
    S => E,P
    E => T_shapes * R_shapes * E0
    P => T_camera * R_camera * P0
    T_shapes => translate(x, y, z)
    R_shapes => rotate(yaw, pitch, roll)
\end{verbatim}
}
For each 3D scene \verb+S+, we perturb the positions and poses of the original cameras (\verb+P0+) and shapes (\verb+E0+) using random translations and rotations.
The position perturbations follow a mixture of uncorrelated Gaussians, and the perturbations 
for pose angles (yaw, pitch \& roll) follow a mixture of von Mises, \ie wrapped Gaussians.
The vector $\beta$ consists of the parameters of the above distributions.

\paragraf{Training setup} Our networks are trained on synthetic images only, and evaluated on NYU Depth V2~\citep{SilbermanECCV2012Indoor} with the same setup as in \citet{ZhangCVPR2016Physically}.
For real images in our optimization pipeline, we sample a subset of images from the standard validation images in NYU Depth V2.
We initialize our network from the synthetically trained model in \citet{ZhangCVPR2016Physically} and
initialize $\beta_0$ using a small value.
To compare with random $\beta$, we construct a dataset of $40$k images with a small random $\beta$ for each image.
We then load the same pre-trained network and train for the same number of 
iterations as in hybrid gradient. We then evaluate the networks on the test set of NYU Depth V2~\citep{SilbermanECCV2012Indoor}, following the same protocol.
The results are reported in Table~\ref{table:nyudepth_all}.
Note that none of these networks has been trained on real images except for validation, and the validation subset of real images is only used to update the decision vector.

\begin{table}[htb]
  \caption{The performance of the finetuned networks on the test set of NYU Depth V2~\cite{SilbermanECCV2012Indoor}, compared to the original network in~\cite{ZhangCVPR2016Physically}. The networks are trained only on the synthetic images. Without optimizing the parameters (random $\beta$), the augmentation hurts the generalization performance. With proper search of $\beta$ using hybrid gradient, we are able to achieve better performance than the original model.}
  \label{table:nyudepth_all}
  \centering
  \begin{adjustbox}{max width=\linewidth}
  \begin{tabular}{lccccc}
    \toprule
    & \multicolumn{3}{c}{Summary Stats $\uparrow$} & \multicolumn{2}{c}{Errors $\downarrow$} \\
    \cmidrule(r){2-4} \cmidrule(l){5-6}
    & $\leq 11.25^\circ$ & $\leq 22.5^\circ$ & $\leq 30^\circ$ & Mean & Median \\
    \midrule
    Original~\citep{ZhangCVPR2016Physically} & $24.1\%$ & $49.7\%$ & $61.5\%$ & $28.8^\circ$ & $22.7^\circ$  \\
    \midrule
    Random $\beta$  +~\citep{ZhangCVPR2016Physically} & $23.0\%$ & $48.8\%$ & $61.3\%$ & $29.2^\circ$ & $23.2^\circ$  \\
    \midrule
    Hybrid gradient +~\citep{ZhangCVPR2016Physically} & $\mathbf{27.3\%}$ & $\mathbf{52.5\%}$ & $\mathbf{63.8\%}$ & $\mathbf{28.1^\circ}$ & $\mathbf{21.1^\circ}$  \\
    \bottomrule
  \end{tabular}
\end{adjustbox}
\end{table}

The numbers indicate that our parametrized generation of SUNCG augmentation exceeds
the original baseline performance.
Note that the network trained with random $\beta$ is worse than original performance.
This means without proper optimization of perturbation parameters,
such random augmentation may hurt generalization, demonstrating that
good choices of these parameters are crucial for generalization to real images.

\subsection{Depth Estimation on Basel Face Model}
\paragraff{Synthetic face generation}
We exploit an off-the-shelf 3DMM morphable face and expression model~\citep{Dai_2017_ICCV,Zhu2015,Zhu2016} to generating human 3D models, with face and pose parameters randomly sampled from mixtures of Gaussians or von Mises. Since the parameters for 3DMM are PCA coefficients, we only include the first 10 principal dimensions each for geometry, texture and expression parameters in the decision vector $\beta$, and uniformly sample for the remaining dimensions to save disk usage.
\paragraf{Training setup} We train a stacked hourglass network~\citep{NewellECCV2016Stacked} from scratch with a single-channel output after a ReLU layer to predict the raw depth, and supervise using mean squared error. The learning rate for the network is $0.1$ and the batch size is $8$.

\begin{figure}[htb]
  \centering
  \includegraphics[width=\linewidth]{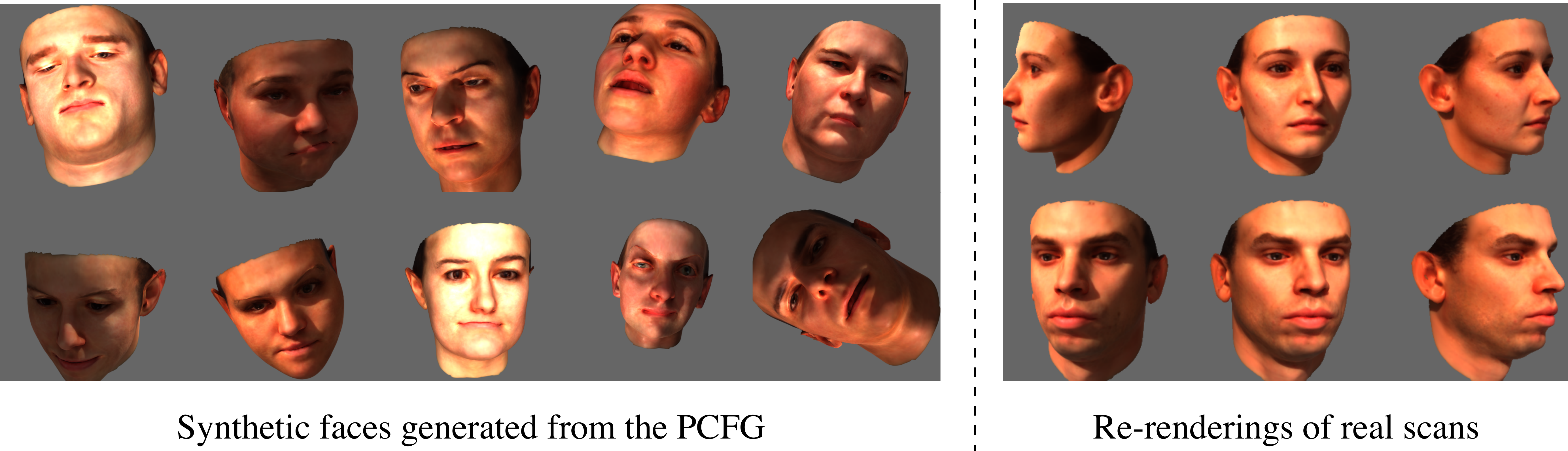}
  \caption{Training images generated using PCFG with 3DMM face model, and example test images.}
  \label{fig:face3d}
\end{figure}

\paragraf{Evaluation} We evaluate on the renderings of the scanned human faces~\citep{Paysan2009}. We split the 10 identities into two disjoint sets for validation and test, then use the rendering parameters provided in the dataset to recreate the renderings as well as depth images. For each scan, there are 3 lighting directions and 9 pose angles, creating $135$ validation images and $135$ test images. Example images are shown in Fig.~\ref{fig:face3d}. For depth evaluation, we use the standard metrics including the relative difference (absolute and squared) and root mean squared error (linear, log and scale-invariant log). The definitions are listed in \citet{NIPS2014_5539} and also detailed in Appendix A.

\begin{table}[hbt]
  \caption{The results on the scanned faces of the Basel Face Model. Our method is able to search for the synthetic face parameters such that the trained network can generalize better.}
  \label{table:bfm}
  \centering
  \begin{adjustbox}{max width=\linewidth}
  \begin{tabular}{lccccc}
    \toprule
    & \multicolumn{2}{c}{Relative Difference} & \multicolumn{3}{c}{RMSE} \\
    \cmidrule(r){2-3} \cmidrule(l){4-6}
    & abs & sqr & linear & log & scale inv. \\
    \midrule
    Random $\beta$ & $0.03718$ & $9.701\times10^{-3}$ & $0.1395$ & $0.1014$ & $0.09717$  \\
    \multirow{2}{*}{\shortstack[l]{Basic Random \\ Search~\citep{HoriaNIPS2018RandomSearch}}} & \multirow{2}{*}{$0.02330$} & \multirow{2}{*}{$1.728\times10^{-3}$} & \multirow{2}{*}{$0.0581$} & \multirow{2}{*}{$0.0299$} & \multirow{2}{*}{$0.02700$} \\
    \\
    \midrule
    Hybrid gradient & $\mathbf{0.02256}$ & $\mathbf{1.649\times10^{-3}}$ & $\mathbf{0.0570}$ & $\mathbf{0.0293}$ & $\mathbf{0.02603}$ \\
    \bottomrule
  \end{tabular}
\end{adjustbox}
\end{table}

The results in \ref{table:bfm} show that our algorithm is able to search for better $\beta$ so that the network trained on the synthetic faces and generalize better on the scanned faces.

\subsection{Intrinsic Image Decomposition on ShapeNet}%
\begin{figure}[htb]
  \centering
  \includegraphics[width=\linewidth]{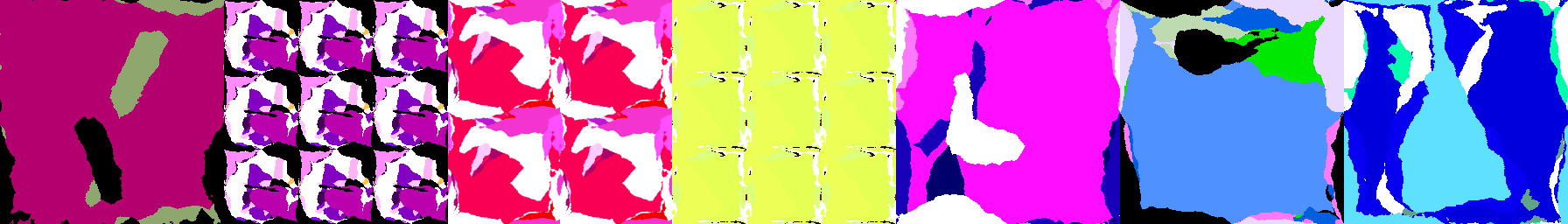}
  \caption{Example textures generated using our procedural pipeline with parameters controlled by $\beta$.}
  \label{fig:csg_tex}
\end{figure}

\paragraff{Texture generation and rendering} We design a painter's algorithm as PCFG for generating the textures. To generate one texture image, we paint Perlin-noise-perturbed polygons sequentially onto a canvas, and then repeat the canvas as the final texture image. The number of repetitions and the number of polygons follow zero-truncated Poisson distributions, the vertex coordinates follow independent truncated Gaussian mixtures, and the number of edges in a polygon are also controlled by sampling probabilities. All the distribution parameters are concatenated to form the decision vector $\beta$. Example textures are shown in Fig.~\ref{fig:csg_tex}.

The texture is then mapped onto the SUNCG shapes~\citep{SongCVPR2017Semantic}. We choose SUNCG shapes because they are well parametrized for texture mapping and we can easily apply our synthetic textures.
We then render the textured shapes using random directional lights as training data. For validation and testing, we randomly render ShapeNet~\citep{Change2015ShapeNet} shapes with their original textures, and randomly choose 50 as validation and 50 for test. The shapes used in validation or test are mutually exclusive.
  \paragraf{Training} We use the Stacked Hourglass Network~\citep{NewellECCV2016Stacked} with a 4-channel output (3 for albedo, 1 for shading), and train with a learning rate of $10^{-4}$ and a batch size of $8$. For supervision, we sum the mean squared error for both albedo and shading outputs as the loss.
  \paragraf{Evaluation}
  We also compare with our Basic Random Search implementation and with the random $\beta$ baseline. We evaluate the performance using mean absolute error (abs), root mean squared error (rmse) and scale-invariant rmse for albedo and shading.
  We also evaluate the reconstruction error of the rendered image, even  though we do not have any supervision for the reconstruction error of the image. The results are shown in Table~\ref{table:shapenetsem}.

\begin{table}[hbt]
  \caption{The results of intrinsic image decomposition on the ShapeNet renderings.}
  \label{table:shapenetsem}
  \centering
  \begin{adjustbox}{max width=\linewidth}
  \begin{tabular}{lcccc}
    \toprule
    & & abs & rmse & rmse (scale inv.) \\
    \midrule
    \multirow{3}{*}{Random $\beta$}
      & Albedo & 0.157 & 0.198 & 0.175 \\
      & Shading & 0.118 & 0.132 & 0.095 \\
      & Reconstruction & 0.139 & 0.169 & -- \\
    \midrule
    \multirow{3}{*}{\shortstack[c]{Basic \\ Random \\ Search~\citep{HoriaNIPS2018RandomSearch}}}
     & Albedo & 0.152 & 0.193 & 0.177 \\
     & Shading & \textbf{0.104} & \textbf{0.116} & \textbf{0.085} \\
     & Reconstruction & 0.134 & 0.166 & -- \\
    \midrule
    \multirow{3}{*}{\shortstack[c]{Hybrid \\ gradient}}
     & Albedo & \textbf{0.147} & \textbf{0.189} & \textbf{0.168} \\
     & Shading & \textbf{0.104} & 0.119 & 0.088 \\
     & Reconstruction & \textbf{0.118} & \textbf{0.150} & -- \\
    \bottomrule
  \end{tabular}
\end{adjustbox}
\end{table}

\section{Conclusion}
In this paper, we have proposed hybrid gradient, a novel approach to the problem of automatically optimizing a generation pipeline of synthetic 3D training data.
We evaluate our approach on the task of estimating surface normal, depth and intrinsic decomposition from a single image.
Our experiments show that our algorithm can outperform the prior state of the art on optimizing the generation of 3D training data, particularly in terms of computational efficiency.

\noindent \textbf{Acknowledgments}
This work is partially supported by the National Science Foundation under Grant No.~1617767.

{\small
\bibliographystyle{ieee_fullname}
\bibliography{main}
}

\newpage
\onecolumn
\renewcommand{\appendixpagename}{Appendix}
\appendixpage
\appendix
\numberwithin{figure}{section}

\section{Metrics}

Here we detail the metrics that we used in the paper. Assume $\mathbf{n}_i$ and $\mathbf{n}^*_i$ are the unit normal vector at $i$-th pixel (of $N$ total) in the prediction and ground truth normal maps, respectively. $d_i$ and $d^*_i$ are depth values of the $i$-th pixel in the prediction and ground truth depth maps, respectively.

\begin{itemize}
\item Mean Angle Error (MAE): $\frac1N\sum_i\arccos(\mathbf{n}_i\cdot\mathbf{n}_i^*)$
\item Median Angle Error (MAE): $\underset{i}{\mathrm{median}}[\arccos(\mathbf{n}_i\cdot\mathbf{n}_i^*)]$
\item Threshold $\delta$: Percentage of $\mathbf{n}_i$ such that $\arccos(\mathbf{n}_i\cdot\mathbf{n}_i^*) \leq \delta$
\item Mean Squared Error (MSE): $\frac1N\sum_i[\arccos(\mathbf{n}_i\cdot\mathbf{n}_i^*)]^2$
\item Absolute Relative Difference: $\frac1N\sum_i |d_i-d_i^*|/d_i^*$
\item Squared Relative Difference: $\frac1N\sum_i (d_i-d_i^*)^2/d_i^*$
\item RMSE (linear): $\sqrt{\frac1N\sum_i(d_i-d_i^*)^2}$
\item RMSE (log): $\sqrt{\frac1N\sum_i(\log d_i-\log d_i^*)^2}$
\item RMSE (log, scale-invariant): $\sqrt{\frac1N\sum_i(\log d_i - \log d_i^*\cdot [\frac1N\sum_i(\log d_i-\log d_i^*)])^2}$
\end{itemize}

\section{MIT-Berkeley Intrinsic Image Dataset}
Our decision vector $\beta$ for PCFG is a 29-d vector, with 4 dimensions representing the probabilities of sampling different primitives, 2 for sampling union or difference, 1 for whether to expand the tree node or replace it with a terminal, 6 for translation mean/variance, 6 for scaling log mean/variance, 2 for sphere radius log mean/variance, 2 for box length mean and variance, 4 for cylinder radius and height log mean/variance, and 2 for tetrahedron length log mean/variance.

\begin{figure}[htb]
  \centering
  \includegraphics[width=\linewidth]{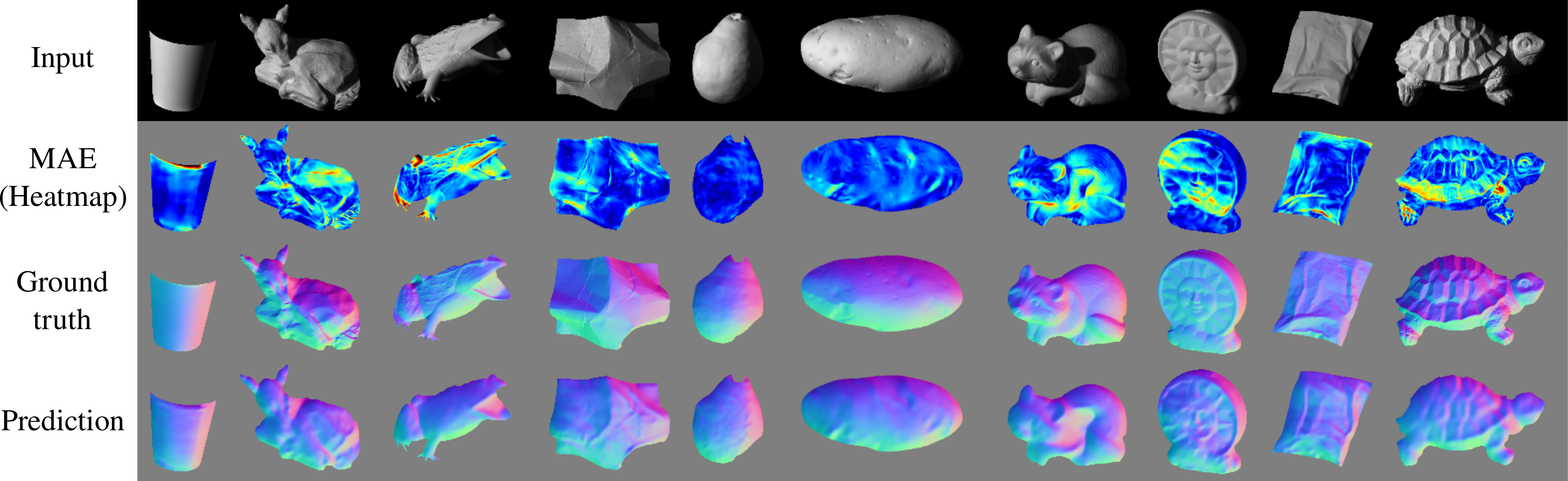}
  \caption{The test set of the MIT-Berkeley Intrinsic Images dataset.}
  \label{fig:mbii_result}
\end{figure}

For optimizing $\beta$, we use the mean angle error loss on the validation set as the generalization loss.
Note that some dimensions of $\beta$ are constrained (such as probability needs to be non-negative), so we simply clip the value of $\beta$ to valid ranges when sampling near $\beta$ for finite difference computation and updating $\beta$. We present the qualitative results in Fig.~\ref{fig:mbii_result}.

\section{NYU Depth V2}
\begin{figure}[hbt]
  \centering
  \includegraphics[width=\linewidth]{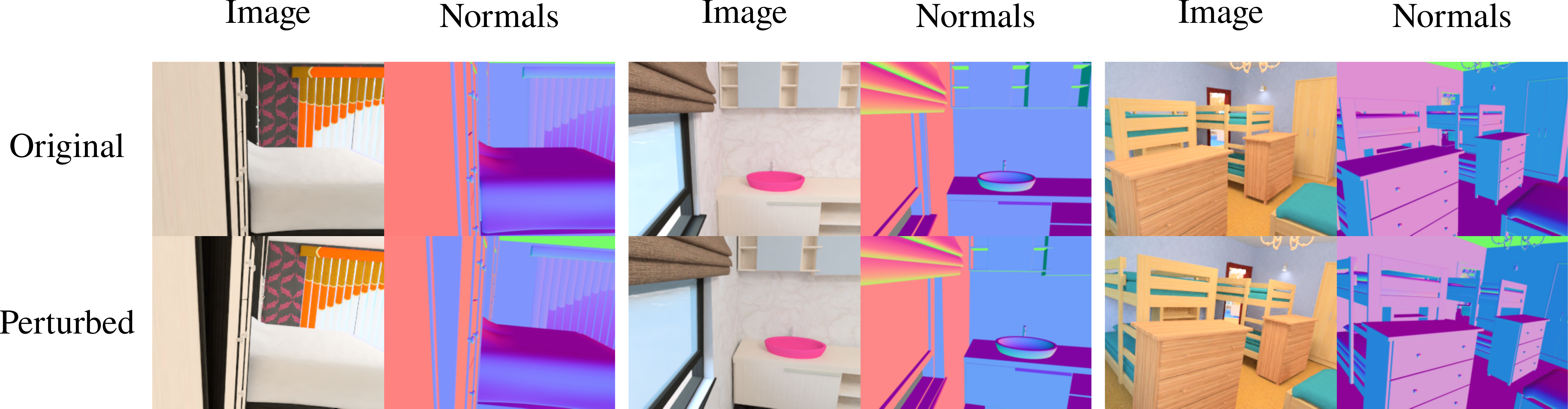}
  \caption{The original scenes in the SUNCG dataset, and our scenes with camera and objects perturbed using our PCFG.}
  \label{fig:nyudepth_vis}
\end{figure}

\begin{figure}[ht]
\begin{minipage}{.33\textwidth}
  \centering
  \includegraphics[width=\linewidth]{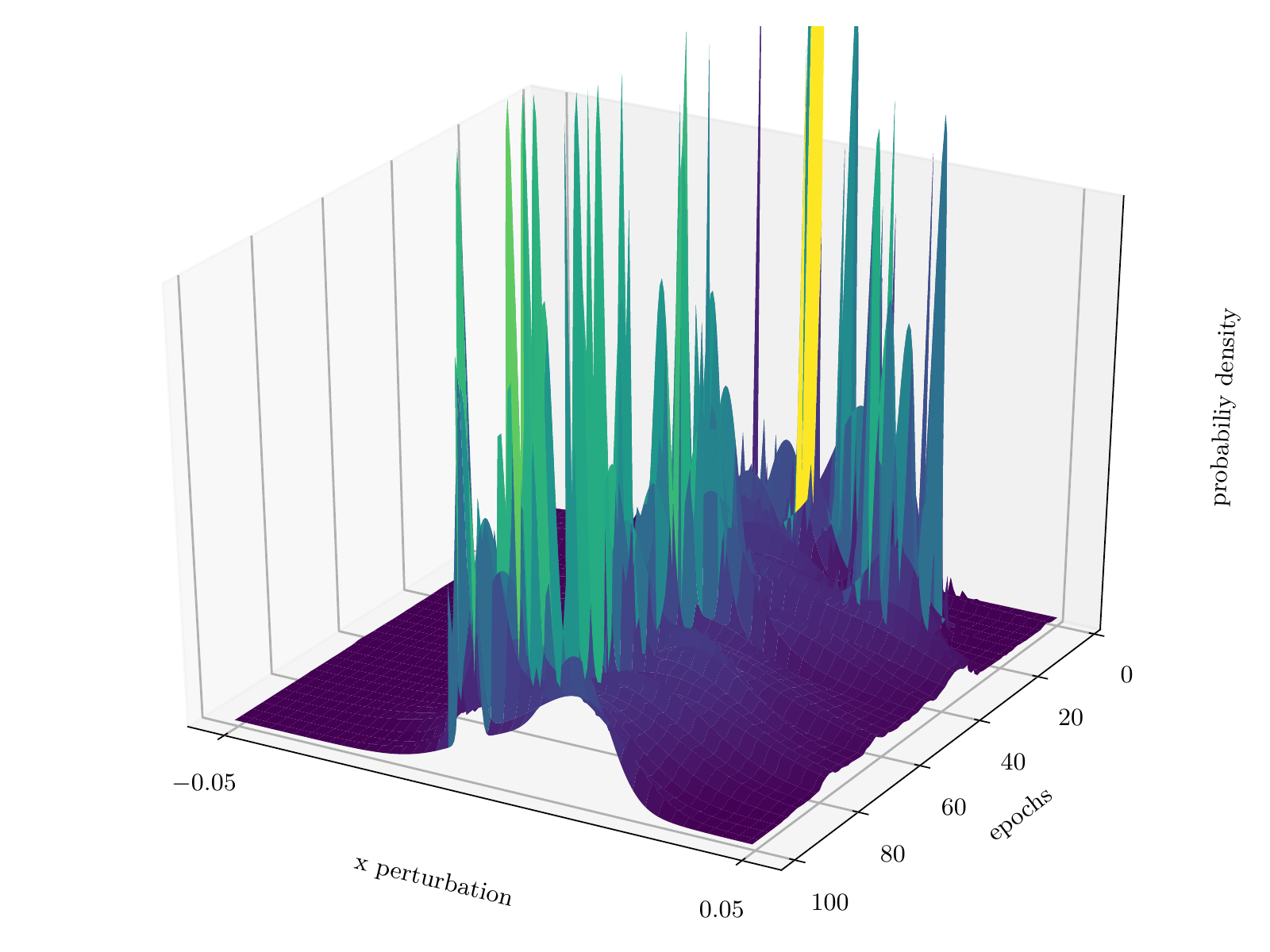}  
\end{minipage}
\begin{minipage}{.33\textwidth}
  \centering
  \includegraphics[width=\linewidth]{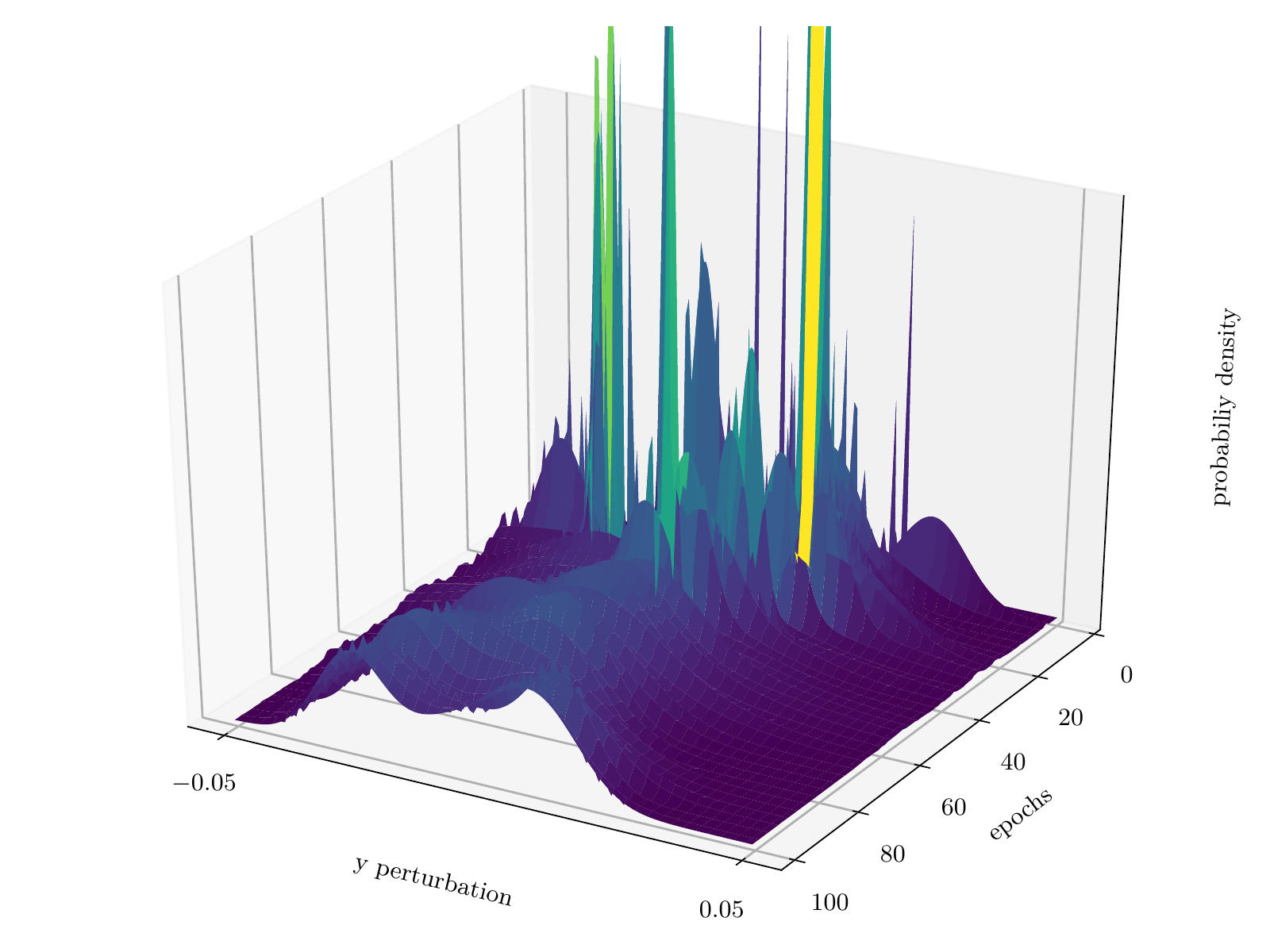}  
\end{minipage}
\begin{minipage}{.33\textwidth}
  \centering
  \includegraphics[width=\linewidth]{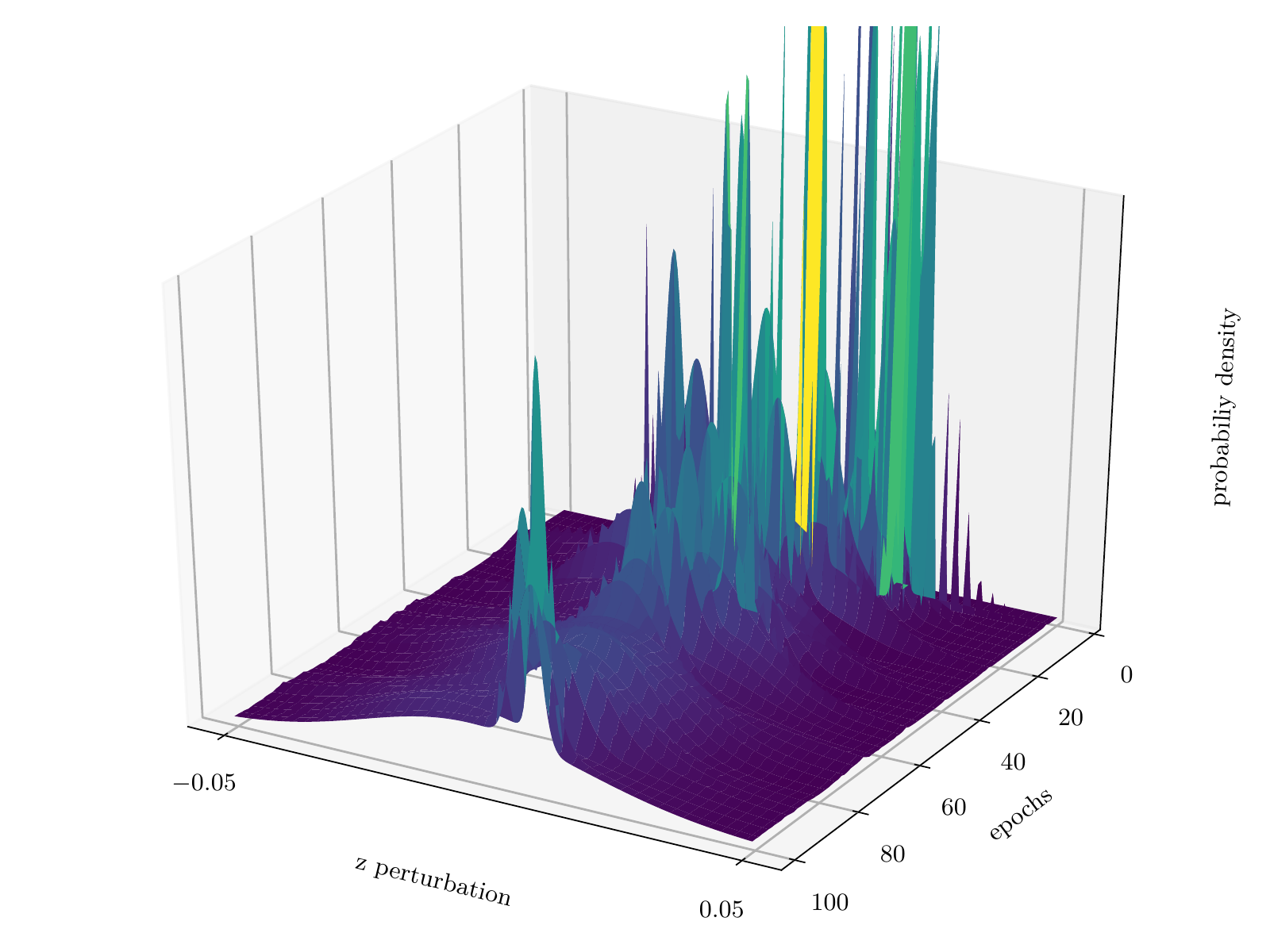}  
\end{minipage}
\caption{How probability distributions change over time for SUNCG perturbation parameters. The three images plot the probability densitity of shape displacement along x, y, z axes respectively.}
\label{fig:dist_change}
\end{figure}

The decision vector $\beta$ is 108-d. It includes the parameters for mixtures of Gaussians/Von Mises for 6 degree-of-freedom (vertical, horizontal and fordinal displacement, yaw, pitch, roll rotation) for shapes in the scene and the camera. Each mixture contains 9 parameters (3 probabilities, 3 means and 3 variances). Examples of perturbed scenes and the original scenes are shown in Fig.~\ref{fig:nyudepth_vis}. The distributions for translation perturbation of shapes are shown in Fig.~\ref{fig:dist_change}.

\section{Basel Face Model}

The decision vector $\beta$ has $204$ dimensions. We use an off-the-shelf 3DMM implementation\footnote{https://github.com/YadiraF/face3d} to generate face meshes and textures for training. The 3DMM has 199 parameters for face identity, 29 for expression and 199 for texture. In implementation, we use 10 principal dimension for face/expression/texture respectively, and randomly sample from a mixture of 3 multivariate Gaussians. Note that the dimensions are independent, so we have a total of $183$ parameters for generating the face mesh. For the 3-dof face pose angle, we also use mixtures of 3 von Mises, which have $21$ parameters in total.

\begin{figure}[htb]
  \centering
  \includegraphics[width=\linewidth]{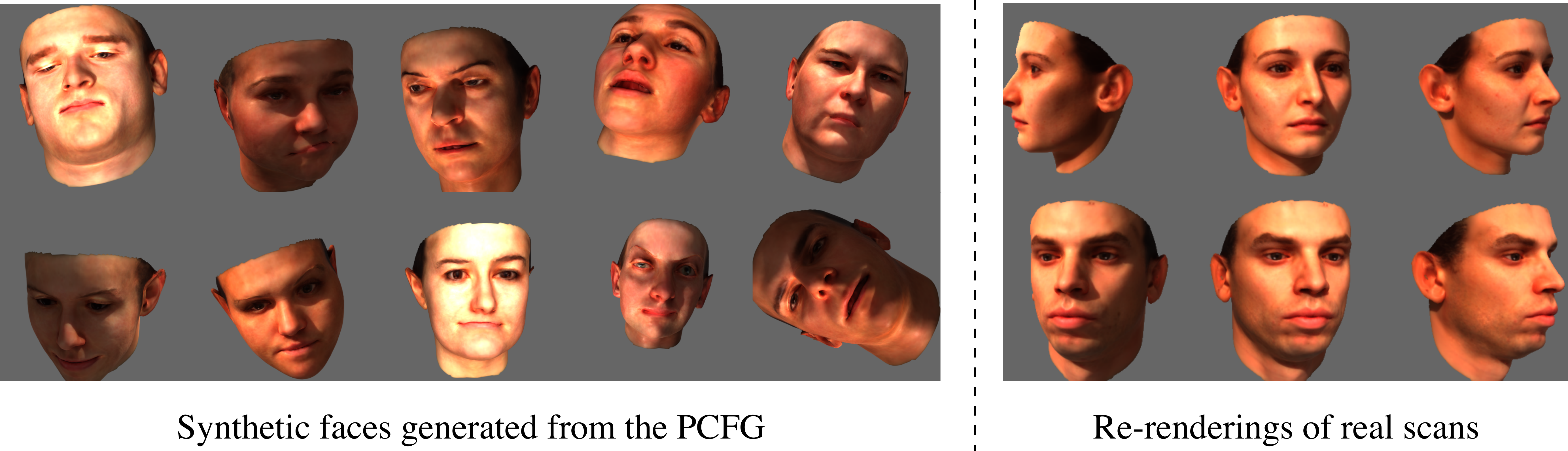}
  \caption{Training images generated using PCFG with 3DMM face model, and 6 example images from the test set.}
  \label{fig:face3d_vis}
\end{figure}

For rendering the training set, we apply a human skin subsurface model using Blender~\citep{Blender2019}, with a random white directional light uniformly distributed on $-z$ hemisphere.
For rendering the test set (the scanned faces in the Basel Face Model), we render with the same 3 lighting angles and 9 pose angles, and the same camera intrinsics as in the original dataset.
Fig.~\ref{fig:face3d_vis} shows the training images randomly generated by the PCFG (left) and example test images(right).

\section{Synthetic Texture Generation for Intrinsic Image Decomposition}
\begin{figure}[htb]
   \centering
   \includegraphics[width=\linewidth]{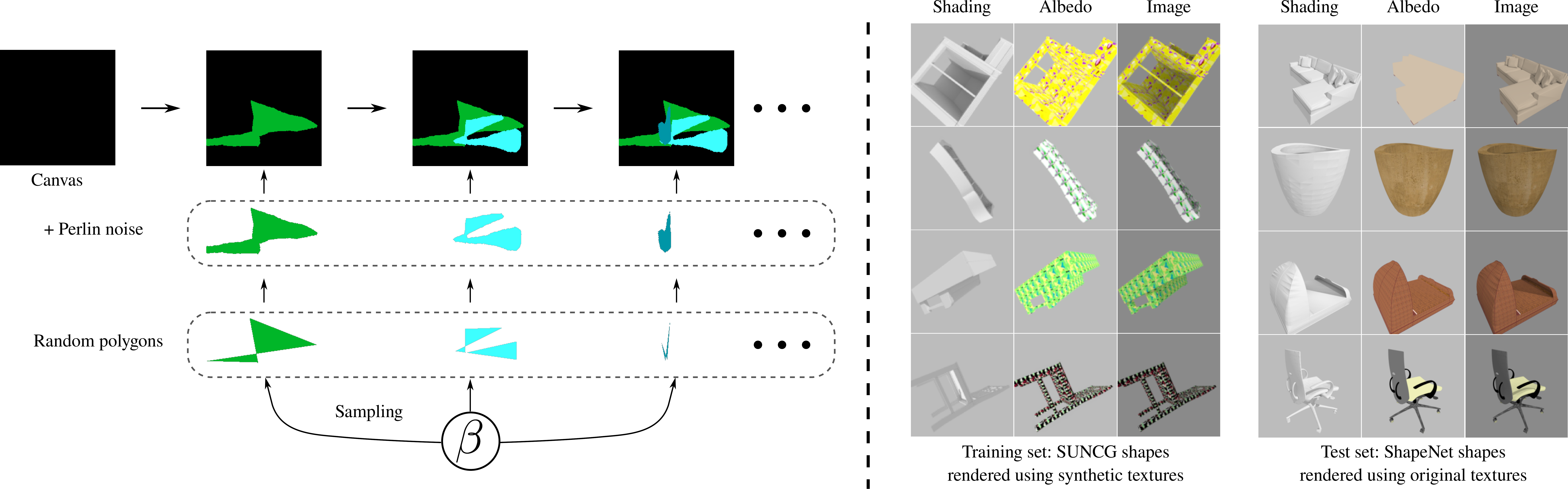}
   \caption{Our texture generation pipeline and example images of the training and test set.}
   \label{fig:texpcfg}
\end{figure}

The decision vector $\beta$ has $36$ dimensions. To sample a texture, we first sample the number of polygons using a zero-truncated Poisson distribution. For each polygon, we then sample the number of vertices (from 3 to 6) according to the probabilities specified in $\beta$. The vertex coordinates of the polygon follow mixed truncated Gaussians.
The polygons are then perturbed using Perlin noise: we first build the signed distance map to the boundary of the polygons, and then perturb the distance using Perlin noise. Finally, we re-compute the boundary according to the distance map to produce the perturbed polygons. These perturbed polygons are then painted onto a canvas to form a texture. This procedure is shown in Fig.~\ref{fig:texpcfg}.

For rendering, we assume the shading is greyscale, and set up a random white directional light. We use Blender~\cite{Blender2019} to render the albedo image and the shading image, then multiple the two images together as the final rendered image. We render SUNCG~\citep{SongCVPR2017Semantic} shapes using our synthetic textures for training, and render ShapeNet~\citep{Change2015ShapeNet} with original textures for validation and test. The examples are also shown in Fig.~\ref{fig:texpcfg}.

\end{document}